\documentclass[10pt, conference]{IEEEtran}
\IEEEoverridecommandlockouts
\usepackage{cite}
\usepackage{amsmath,amssymb,amsfonts}
\usepackage{booktabs}
\usepackage{array}
\usepackage{tabulary}       % content-balanced column widths for text-heavy tables
\usepackage{siunitx}        % decimal-aligned numeric columns
\usepackage{threeparttable} % clean notes set under the table, not in the caption
\usepackage{graphicx}
\usepackage{xcolor}
\usepackage{textcomp}
\extrafloats{48}                     % headroom for the full-width floats
\newcommand{\ind}[1]{\mathbb{1}\!\left[#1\right]}

\begin{document}

% Apply one bibliography-wide author-list policy: entries with more than six
% authors show the first author followed by "et al." while refs.bib retains
% complete author metadata where available.
\bstctlcite{IEEEtranBSTcontrol}

%\title{Locating Computation, Perception, and Agency in\\ Tool-Augmented Time-Series Question Answering}

\title{T-SMART: Mechanism-Level Attribution for Tool-Augmented Time-Series Question Answering}

%\author{
%\IEEEauthorblockN{Anonymous}
%\IEEEauthorblockA{}
%}

\author{
    \IEEEauthorblockN{
        Ivan Delgado\textsuperscript{1},
        Himansi Gupta\textsuperscript{1}, 
        Bishal Khatri\textsuperscript{1}, 
        Niharika Sapre\textsuperscript{1},
        Lameta Shamoon\textsuperscript{1}, 
        Onat Gungor\textsuperscript{2},
        Tajana Rosing\textsuperscript{1}
    }
    \IEEEauthorblockA{
        \textsuperscript{1}University of California, San Diego, CA, USA \\
        \textsuperscript{2}West Virginia University, Morgantown, WV, USA \\
        \{ivdelgado, h4gupta, bikhatri, nsapre, lashamoon, tajana\}@ucsd.edu, onat.gungor@wvu.edu
    }
}

\maketitle

\begin{abstract}
Large language models (LLMs) can struggle with time-series question answering (TS-QA), especially when numerical signals are serialized as text and require explicit computation. Tool-augmented approaches improve performance, but existing systems often intertwine language reasoning, computation, and perception, making it difficult to determine which components drive the gains. We present \textsc{T-SMART}, a neurosymbolic framework that separates these roles: a frozen LLM interprets questions and selects operations, deterministic tools perform numerical computation, and structured perception is invoked only when needed. Controlled paired ablations show that deterministic computation provides the dominant benefit, improving accuracy by 31.7 percentage points over direct LLM reasoning on serialized time series, while language understanding and perception offer smaller complementary gains. These results indicate that tool-augmented TS-QA benefits primarily from reliable numerical execution rather than additional language-model reasoning, and provide a controlled framework for analyzing component contributions in neurosymbolic time-series systems.
\end{abstract}


\section{Introduction}\label{sec:intro}
Time series underpin decision-making in finance, energy, and healthcare, where many questions reduce to recurring tasks such as detecting trends, periodicity, anomalies, or correlations~\cite{gungor2024robust}. LLMs offer a flexible interface for such queries, but direct reasoning over numerical time series remains unreliable. Merrill et al.~\cite{merrill24} find near-chance zero-shot performance and greater reliance on linguistic than numerical cues, while serialization increases context length and fragments numerical values into unstable token sequences \cite{gruver23}. Related studies show that removing the language-model component from forecasting pipelines may not hurt performance \cite{tan24}, and recent benchmarks report that leading commercial models remain below two-thirds accuracy on TSAQA and ARFBench \cite{jing26,xie26}. These findings suggest that stronger language models alone may be insufficient for reliable time-series reasoning.

A growing body of work addresses these limitations through tool-augmented and agentic approaches. TS-Agent~\cite{liu25}, for example, keeps time-series data in numerical form and uses an LLM to orchestrate analytical tools within a ReAct loop, while TimeART~\cite{wu26} trains an agent to select and invoke tools from a large analytical corpus. More broadly, recent work spans increasingly complex reasoning strategies, from direct model inference to branch-structured exploration~\cite{chang26}. These systems show that external computation can substantially improve time-series reasoning. However, because language interpretation, tool selection, numerical computation, and perception are often tightly integrated, it remains unclear which components are primarily responsible for the observed gains.

This paper asks a central question: in tool-augmented TS-QA, which components are actually responsible for performance gains? We investigate this question with \textsc{T-SMART}, a single-pass, instrumented framework designed to isolate the contribution of each component. Rather than adding increasingly complex agentic behavior, \textsc{T-SMART} separates language interpretation, deterministic computation, conditional perception, and answer verification so that each can be evaluated independently. Our comparisons hold the language-model backbone, prompts, and evaluation examples fixed, and our claims are restricted to benchmark questions supported by the registered tool set. This controlled design allows us to measure where performance improvements originate. In particular, we find that deterministic computation provides the dominant gain, while additional evidence-layer correction offers no measurable improvement once verified tool outputs are available. Our contributions are as follows.

\begin{enumerate}
    \item We introduce a mechanism-level decomposition of tool-augmented TS-QA that isolates the contributions of language understanding, deterministic computation, conditional perception, and evidence-layer correction.

    \item We distinguish architectural gains from backbone effects. At a fixed language-model backbone, the proposed tool architecture improves accuracy by 13.5 percentage points over direct model reasoning, enabling a controlled estimate of the benefit of tool augmentation.

    \item We evaluate conditional perception by comparing structured visual inputs with raw pixels across two replications, measuring both accuracy and susceptibility to misleading visual cues.

    \item We evaluate three post-tool evidence-correction mechanisms using paired tests, trigger rates, and power analysis. Two never activate, while the third does not significantly improve accuracy.

\end{enumerate}

\begin{figure}[t]
\centering
\includegraphics[width=0.75\linewidth]{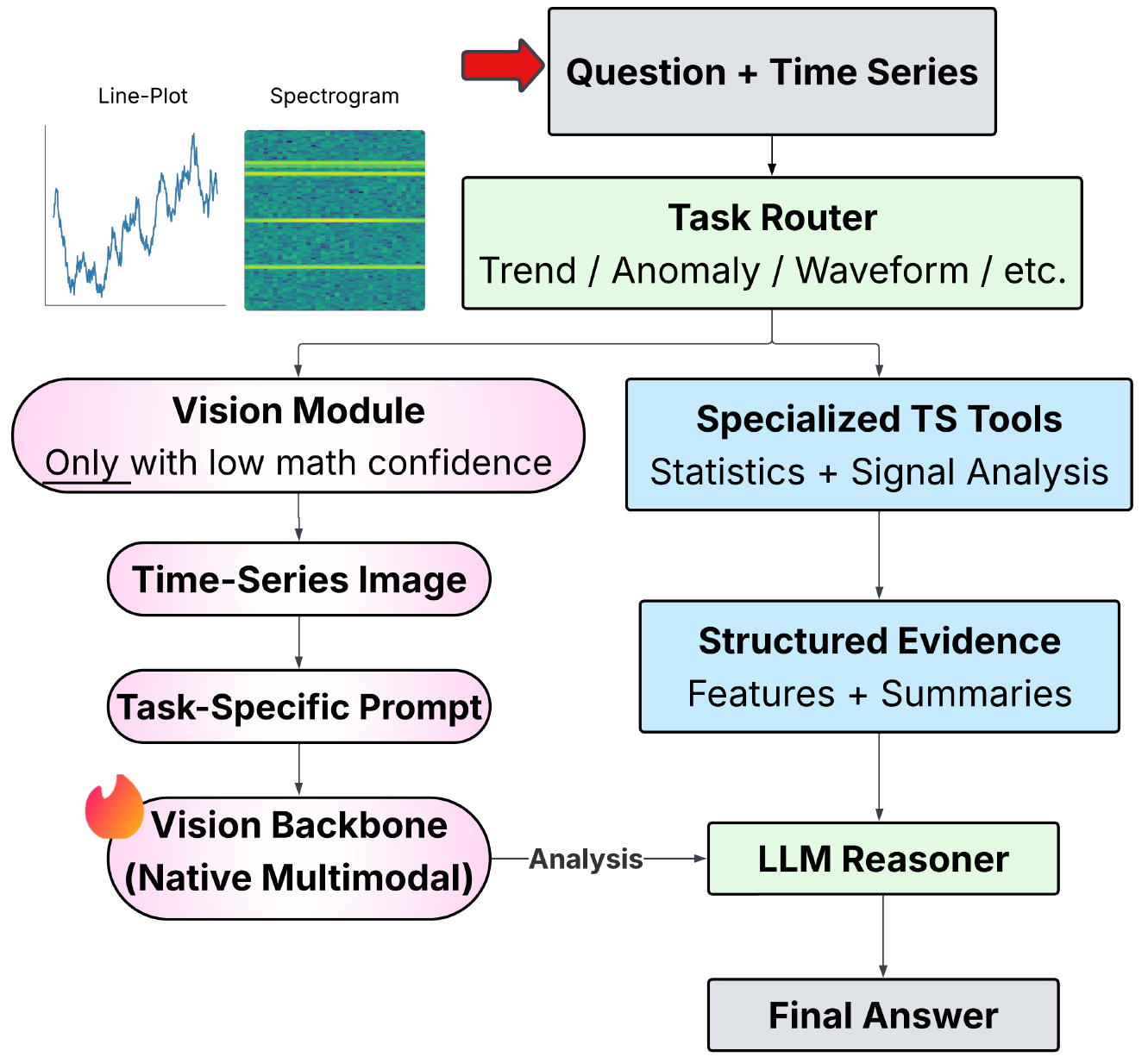}
\caption{The single-pass T-SMART pipeline. One frozen model serves as router, vision sensor, and answer reader; deterministic tools perform every numerical computation, and the visual sensor is invoked only when the trust gate fires.}
\label{fig:pipeline}
\end{figure}

\section{Related Work}\label{sec:related}
Early work applied language models to time series by serializing numerical sequences as text for forecasting \cite{gruver23}, while later evaluations exposed the limitations of this approach for numerical reasoning \cite{merrill24,tan24}. A natural response is to move computation outside the language model, following tool-augmented methods that delegate explicit calculations to external programs \cite{schick23,gao23,chen23pot,yao23}. Sprague et al. further show that chain-of-thought gains are strongest on symbolic execution tasks, where dedicated solvers can outperform prompted reasoning \cite{sprague24}. We examine if this separation between language reasoning and computation benefits time-series question answering.

Several benchmarks quantify the difficulty of time-series QA. TimeSeriesExam \cite{cai24} evaluates five reasoning categories, while MMTS-Bench \cite{yin26} introduces out-of-distribution structure across four subsets and combines multiple-choice and numerical questions. It shows that time-series-adapted LLMs can lag general-purpose models across domains. Broader and more challenging benchmarks corroborate the performance gap discussed in Section~\ref{sec:intro} \cite{yu26,jing26,xie26}, while others extend evaluation to multi-turn, open-ended, and cross-modal settings \cite{kong26mt,kong25,chen25}. Statistical significance testing remains uncommon: ARFBench reports bootstrap confidence intervals, but paired tests between systems are generally absent.

A second line of work explores tool- and agent-based approaches. For time series, TS-Agent \cite{liu25} and TimeART \cite{wu26} adopt tool-grounded reasoning with explicit evidence and tool use. TS-Agent ablates components such as its critic and answer verification, but does not isolate iterative reasoning from a single pass using the same tools; to our knowledge, we provide the first such comparison. Kim et al.\ take a complementary non-agentic approach by compiling the time series into a deterministic structured program before model reasoning \cite{kim26}. Under the taxonomy of Chang et al.~\cite{chang26}, T-SMART falls within the direct-reasoning family: its router selects one tool branch rather than exploring or aggregating multiple reasoning paths. Thus, T-SMART shares the tool-grounding premise of prior work but limits autonomy to a single route and evidence pass, treating iteration as a hypothesis to be tested.

Self-consistency improves chain-of-thought by sampling and voting \cite{wang22}; it is a decoding baseline, with no revision of an answer against feedback. Huang et al.\ show that intrinsic self-correction frequently fails without an external signal, while naming tool-verified correction as a regime their result leaves open \cite{huang24}. Our loops operate in that open regime, correction driven by tool evidence, so the null of Section~\ref{subsec:nulls} is new evidence rather than a replication.

The evidence for vision is mixed. Rendering a series as an image can assist a vision-language model \cite{daswani24}, yet that study's own per-task results concentrate the benefit on perception tasks and reverse on others, and multimodal fusion for time series frequently underperforms \cite{yu26}. Conditional invocation has precedent outside time series in AutoTool, which learns by reinforcement when a visual tool call is useful \cite{ma26}; our gate is instead triggered by deterministic statistical evidence, a coarse, recall-oriented rule rather than a calibrated selector (Section~\ref{subsec:gate}). The structured half of the design belongs to the chart-derendering literature, where DePlot translates a plot to a table before a language model reasons over it \cite{deplot}; our sensor differs in being invoked conditionally, in emitting a topology summary rather than a data table, and in being evaluated against a raw-pixel condition on matched rows.

Evaluation methodology affects the interpretation of any reported gain. Saqur et al.~\cite{saqur26} argue that reported gains in time-series modeling are often illusory once evaluation is taxonomy-aware. Their argument concerns forecasting; we adopt the same caution for TS-QA and report matched-backbone paired differences as primary evidence, with cross-system tables given for context.

\section{Method}\label{sec:method}

\begin{table}[t]
\centering
\caption{Deterministic branches and the core tools each invokes; the per-branch subtype registry is released with the code.}
\label{tab:branches}
\footnotesize
\begin{tabulary}{\linewidth}{@{}l L@{}}
\toprule
Branch & Core tools \\
\midrule
trend & OLS slope, functional-form fits, \texttt{ruptures} change-points \\
periodicity & FFT and spectral features, seasonal decomposition \\
anomaly & z-score and IQR scoring, change-point detection \\
noise & Engle ARCH-LM, Ljung-Box, ADF, KPSS \\
similarity & dynamic time warping, statistical distances \\
causality & Granger F-test, lag correlation \\
\bottomrule
\end{tabulary}
\end{table}

\subsection{Problem and design rationale}\label{subsec:problem}
A TS-QA instance is a triple $x = (Q, \mathbf{X}, O)$: a natural-language question $Q$, one or more numerical series $\mathbf{X}$, and a possibly empty option set $O$. A deterministic rule over the question and options, independent of any model output, assigns each instance an answer schema
\begin{equation}\label{eq:schema}
\sigma(Q, O) \in \{\texttt{mcq},\ \texttt{numerical},\ \texttt{categorical}\}.
\end{equation}
On \texttt{mcq} rows the system returns a choice $\hat{a} \in O$, scored by exact match. The \texttt{categorical} value marks rows that request a free-text label; no scored row in either benchmark takes it. On \texttt{numerical} rows the system returns a value $\hat{v} \in \mathbb{R}$, scored at a relative tolerance,
\begin{equation}\label{eq:acc10}
\mathrm{Acc\text{@}10}(\hat{v}, v^{*}) = \ind{\,\lvert \hat{v} - v^{*} \rvert \le 0.1\,\lvert v^{*} \rvert\,}.
\end{equation}

Because language models process $\mathbf{X}$ as text and lose precision \cite{merrill24,gruver23}, T-SMART treats the limitation as a fixed constraint, routing every numerical computation to deterministic code and assigning the model two functions: selecting the computation and expressing its result as an answer.

\subsection{The pipeline}\label{subsec:pipeline}
T-SMART is a single-pass decision procedure without a loop (Figure~\ref{fig:pipeline}). One frozen model fills three roles, router, vision sensor, and answer reader, at temperature 0, and is never fine-tuned. The bounded loops of Section~\ref{subsec:nulls} are extensions; none is enabled in the reported default.

The router runs first. It reads $Q$ and a flag for the number of series present and returns a JSON route (branch, scope, subtype, series type) without observing the answer options. The deterministic schema rule of Eq.~\eqref{eq:schema} then runs, and the pipeline splits on its output.

On a multiple-choice row, the routed branch runs deterministic estimators on $\mathbf{X}$ and writes a structured evidence dictionary; the verifier raises flags and evaluates the trust gate; the answer reader returns a letter, preceded by the structured vision sensor when the gate fires. On a free-response row, control exits at the numeric head, before the gate and the reader: the head parses the question into an operation plan over a closed registry and computes the value with deterministic code; there the only model call is the router, and the vision sensor and answer reader are never invoked.

A routing failure, a disabled branch, or a row without a series falls back to a direct prompt rather than aborting the run.

\subsection{Deterministic branches}\label{subsec:branches}
Each branch runs a fixed set of estimators (Table~\ref{tab:branches}). Causality is the clearest case: every model in TimeSeriesExam struggles most there \cite{cai24}, and it is the category where a Granger test offers the clearest advantage.

The bank is designed to mirror the category structure of the benchmarks it is evaluated on: the claims in this paper concern taxonomy-covered TS-QA, the questions that reduce to a computation the bank registers, and Section~\ref{subsec:main} reports performance outside that coverage, where rows fall back to a direct prompt.

\subsection{Verifier and the trust gate}\label{subsec:gate}
The verifier applies rule-based checks to the evidence dictionary and raises a set $F$ of flags: weak trend fits, unreliable spectral periods, disagreeing stationarity tests, non-significant Granger results, weak correlations, detected ARCH effects, high volatility, and incomplete evidence. The flag \texttt{evidence\_incomplete} additionally routes the row toward fallback.

The gate computes a trust score over the flags and a second set $D(e, q)$ of graded diagnostics derived from the evidence $e$ and the question $q$: ambiguous functional-form fits, marginal Granger $p$-values, high ADF $p$-values, correlation-versus-Granger conflicts, unstable chunk statistics, and weak seasonal, trend, or anomaly signals on questions that ask about them. Each raised flag deducts a fixed 0.15 and each diagnostic deducts its weight $w_d \in [0.05, 0.25]$,
\begin{equation}\label{eq:trust}
s = \max\Bigl(0,\; 1 - 0.15\,\lvert F \rvert - \sum_{d \in D(e,q)} w_d\Bigr),
\end{equation}
and the gate fires according to
\begin{equation}\label{eq:trigger}
g(e, q) =
\begin{cases}
1, & \mathrm{vis}(q), \\
0, & \mathrm{HC}(e) \text{ and not } \mathrm{vis}(q), \\
1, & s \le 0.85, \\
0, & \text{otherwise},
\end{cases}
\end{equation}
evaluated top to bottom, where $\mathrm{vis}(q)$ marks an explicitly visual question and $\mathrm{HC}(e)$ is a high-confidence predicate (strong fit margins, decisive Granger evidence, or strong seasonal or trend strength, available only when no flag is raised) that vetoes the trigger. The exact flag definitions, diagnostic weights, predicate, and render-selection logic are released with our code, and two configuration switches override the gate after it runs.

Two properties of this rule matter for interpreting the results: the gate is permissive by construction, since a single raised flag deducts 0.15 and satisfies $s \le 0.85$ on its own, and its firing rate is correspondingly high, 568 of 746 TimeSeriesExam rows (76.1\%). We designed the gate for recall and claim no selectivity for it; the benefit of gating is instead measured against counterfactuals, a math-only configuration and a learned suppression policy that lowers the look-rate to 27.9\% (Sections~\ref{subsec:config} and~\ref{subsec:audits}). All constants, including the agency quality threshold (0.55), were fixed before the reported paired evaluations; none was learned or statistically identified.

\subsection{Conditional structured vision and the SVI}\label{subsec:vision}
When the gate fires, the pipeline renders $\mathbf{X}$ as a spectrogram for periodicity, noise, causality, or spectral flags, and as a line plot for trend, shape, and the waveform subtypes. The vision sensor reads the image and returns a structured JSON description (topology, waveform, breakpoints). Only that description enters the answer step; the raw pixels do not. We call this property the Structured-Vision Invariant (SVI).

We treat vision as one more tool, called when the statistics are judged insufficient and otherwise withheld, because a rendered series can mislead as readily as it can inform: the mere presence of an image biases the answer model toward reporting a pattern even when the statistics reject one, a failure we call visual sycophancy. The term's one prior use concerns answer shifts toward an option pre-marked in the image itself \cite{lim24}, and related work studies VLM sycophancy under leading textual prompts \cite{zhao24}; the effect here requires neither: a neutral render alone induces answers that contradict the statistics. Sycophancy is the reason perception is gated. We measure whether the structured reading also reduces sycophancy, and the result is inconclusive; the claim we make for the SVI is auditability, since the answer step consumes a verified, inspectable reading rather than pixels, and its accuracy cost is reported.
The design choice is contested in the literature \cite{daswani24,yu26}, so we measure it. A raw-pixel ablation attaches the rendered PNG directly to the answer model, compared on matched rows against the structured sensor and against the math-only counterfactual. The comparison has been run twice on the TimeSeriesExam gated subset ($n = 568$), an initial run and a pinned-commit regeneration; Section~\ref{subsec:mechanism} reports both replications, the cost of the invariant, and the sycophancy rates under each visual condition.

\subsection{Numeric head}\label{subsec:numhead}
TimeSeriesExam is multiple choice. MMTS-Bench is approximately half free-response on its Base subset, with questions that request a number. The numeric head answers these.

The head is gated by the answer schema of Eq.~\eqref{eq:schema}, computed before any model call: a row with two or more parseable options is classified \texttt{mcq} and cannot enter the head, so every multiple-choice result is byte-identical with the head enabled or disabled. An audit on the 686 labelled Base rows found a single off-diagonal row (a multiple-choice row inferred as numerical) and no numerical row misread as multiple choice, so the byte-identity property rests on a measured 0/388 (Section~\ref{subsec:audits}). The property concerns enabling or disabling the head within the default pipeline; \texttt{llm\_compute} is a different, tool-free condition (Section~\ref{subsec:config}).

A parser converts the question into a plan over a closed registry $\Pi$: a base quantity (\texttt{std}, \texttt{mean}, \texttt{median}, \texttt{range}, and others) optionally wrapped in a compose operation, with numeric parameters where required. The invariant is on the output: the planner emits a validated operation plan $\pi \in \Pi$ and never the final value, which deterministic code then computes,
\begin{equation}\label{eq:head}
\hat{v} = \mathrm{eval}(\pi, \mathbf{X}), \qquad \pi = \mathrm{parse}(Q) \in \Pi.
\end{equation}
Where PAL and Program-of-Thoughts have the model emit arbitrary programs \cite{gao23,chen23pot}, the head restricts it to this closed registry, a restriction chosen for auditability. We study three parsers as an experimental axis: \texttt{A0}, a keyword parser; \texttt{A1}, a deterministic composition grammar with fuzzy and typo matching; and \texttt{A2}, an LLM planner that fires only when \texttt{A1} abstains and still emits a validated plan. Free-response answers are scored by Eq.~\eqref{eq:acc10}.

\subsection{Configuration as the unit of experiment}\label{subsec:config}
Behavior is changed by switching pipeline arguments, never by editing prompts toward an answer: a configuration is a named dictionary of those arguments, prompts are identical across configurations, and every experiment is a paired comparison of two configurations on the same rows. Three schema-conditioned axes define a configuration: vision policy acts only on rows that reach the visual gate (the multiple-choice rows), the numeric head acts only on free-response rows (the \texttt{A0} default, \texttt{A1}/\texttt{A2} parser variants, or a bypass), and agency is disabled by default.

The canonical system, \texttt{baseline}, is therefore schema-conditioned: on multiple-choice rows it is the trust gate with no loop and no numeric head; on free-response rows it is the \texttt{A0} numeric head, reached before the gate. Across evaluation harnesses the behavior-changing switches are copied unchanged and each harness adds inert defaults, so configurations are semantically identical rather than byte-identical dictionaries. Table~\ref{tab:configs} lists the experimental conditions.

\begin{table}[t]
\centering
\caption{Configurations used as experimental conditions. A configuration is a named set of pipeline arguments; prompts are identical across conditions.}
\label{tab:configs}
\footnotesize
\begin{threeparttable}
\begin{tabulary}{\linewidth}{@{}l L@{}}
\toprule
Name & Setting and role \\
\midrule
\texttt{baseline} & trust gate, no loop; \texttt{A0} numeric head on free-response. Canonical T-SMART; the control condition \\
\texttt{vision\_override} & suppress vision on noise, force anomaly line plots. TimeSeriesExam-selected routing intervention, frozen before OOD testing; a mechanism-transfer ablation \\
\texttt{learned\_gate} & per-branch suppression filter over the trust gate; a lower-look-rate policy fitted on MMTS discordance, evaluated on held-out TimeSeriesExam \\
\texttt{vision\_off} & suppress vision on all branches; the math-only counterfactual \\
\texttt{head\_grammar}, \texttt{head\_planner} & \texttt{A1} / \texttt{A2} parser over the \texttt{A0} default; free-response numeric-head variants \\
\texttt{llm\_only} & no router, branch, or tools; one direct answer call. Raw-model control for the fixed-backbone analysis \\
\texttt{llm\_compute} & tool-free on every row: the model computes the value from the raw series on numerical rows; multiple-choice rows take the direct \texttt{llm\_only} path. Numeric-head bypass counterfactual \\
\bottomrule
\end{tabulary}
\begin{tablenotes}[flushleft]\footnotesize
\item The names shown are those used in this paper; the released code registry keeps its original identifiers, and the mapping is documented in the release.
\end{tablenotes}
\end{threeparttable}
\end{table}

The \texttt{vision\_override} condition was selected on TimeSeriesExam as a narrow routing intervention and frozen before any MMTS evaluation (Table~\ref{tab:configs}). The \texttt{learned\_gate} condition is a per-branch suppression filter trained only on rows where the trust gate fired, able to suppress vision calls but never add one; its held-out result is reported in Section~\ref{subsec:audits}.

\subsection{Backbones}\label{subsec:backbones}
The main reported T-SMART results use gemini-3.1-flash-lite at temperature 0, with that single model filling all three roles; a secondary analysis uses gpt-4o-mini, the model TS-Agent reports \cite{liu25}, to separate architecture from backbone at a fixed model. Models are never mixed within a run.

\subsection{Evaluation}\label{subsec:eval}
We evaluate on two benchmarks. TimeSeriesExam \cite{cai24}, 746 multiple-choice questions across five categories, is the primary benchmark; MMTS-Bench \cite{yin26}, with four subsets, is the out-of-distribution benchmark, where Base free-response is scored by Eq.~\eqref{eq:acc10} and multiple choice by exact match.

The protocol is paired throughout \cite{dror18}. Each comparison's decision rule, its hypotheses, primary contrast, and stratification, is committed to a repository registration file with the exact command before scoring; no external registry is used, so we say specified before scoring rather than pre-registered, and the files, with any deviations, are part of the release. Both conditions run on the same rows, and the reported estimate is the mean paired difference $\hat{\Delta} = \tfrac{1}{n} \sum_{i} \bigl(c_i^{\mathrm{T}} - c_i^{\mathrm{C}}\bigr)$, where $c_i \in \{0, 1\}$ is per-row correctness under the treatment and control conditions. The 95\% confidence interval is a percentile bootstrap over the paired per-row differences, with the row as the resampling unit, 10{,}000 resamples, and a fixed seed. Rows generated from the same template or series are treated as independent. That simplification matters for results near a decision threshold, notably the \texttt{vision\_override} overall difference ($p = 0.052$) and the propose-loop null ($p = 0.79$): we flag both as sensitive to the independence assumption, and the released tooling supports a per-series cluster bootstrap for the check. Significance uses the exact conditional McNemar test, conservative relative to mid-$p$ variants \cite{fagerland13}, on the discordant counts $n_{10}$ (treatment-only correct) and $n_{01}$ (control-only correct). We stratify by the pre-reroute router branch (\texttt{initial\_branch\_used}), because an agency loop can rewrite the post-hoc branch and conditioning on it would condition on a collider. The per-branch family is corrected with Holm, and parsing errors, tool failures, and timeouts all count as incorrect.

We do not claim bit-exact runs, and we measure routing stability directly: every condition issues its own router call on every row, no route is cached or reused across conditions, and for each paired comparison we report routing drift, the fraction of paired rows whose routed branch differs. On the gemini-3.1 main runs the measured drift was 0 of 746 rows in every comparison, so the configuration was effectively the only varying factor. That figure is a per-row equality check on the routed branch, run for each comparison; the same check exposes the instability of the other backbone. The gpt-4o-mini router disagrees with itself on approximately 40\% of rows at temperature 0, with flips concentrated in and out of the fallback route (the anomaly-routed stratum's size swings between 49 and 71 rows across identical runs); API-level nondeterminism at temperature 0 is documented behavior \cite{ouyang23}. Its consequence for inference is that gpt-4o-mini runs are treated as stress tests: per-branch differences under that router are not read as mechanism evidence, and conclusions there rest on the discordant-pair McNemar test and a held-out replication. A route-pinned design, caching one set of routes and replaying them in both conditions, would remove this noise at the cost of evaluating a counterfactual router; we did not run it, and mark these analyses secondary. Primary evidence throughout is the matched-backbone paired difference \cite{saqur26}.

\subsection{Tested extensions: bounded evidence-layer correction}\label{subsec:nulls}
We implemented three bounded correction loops, none enabled in \texttt{baseline}. All three correct in the evidence layer; none re-prompts the answer model. This axis, bounded evidence-layer correction, is the agency this paper tests, a narrower regime than full agentic control.

The first, reroute-once, re-routes a single time on an incompleteness flag. The second, refine, accumulates a second branch's evidence below a quality threshold using keyword candidate selection and no additional model call. The third, propose, selects the next tool from a closed registry at one model call per step. Trigger rates and paired effects are reported in Section~\ref{subsec:mechanism}.

The loops operate in the tool-verified regime that Huang et al.\ leave open \cite{huang24}; majority-vote self-consistency \cite{wang22} serves as a decoding-time control.

\section{Experiments}\label{sec:exp}

\subsection{Setup}\label{subsec:setup}
All main T-SMART runs use gemini-3.1-flash-lite at temperature 0, on the full 746-row TimeSeriesExam and the four MMTS-Bench subsets, with every comparison paired on identical rows under the statistics of Section~\ref{subsec:eval}.

\subsection{Main accuracy and comparison}\label{subsec:main}
Tables~\ref{tab:tsexam} and~\ref{tab:mmts} compare T-SMART with published systems on TimeSeriesExam and MMTS-Bench. Both tables follow the scoping of Section~\ref{subsec:branches}: the tool bank targets these benchmarks' taxonomy, and the InWild and Align columns of Table~\ref{tab:mmts} are the off-taxonomy exception. Cross-system comparisons are reported for context only: the T-SMART rows use gemini-3.1-flash-lite on our harness, while the baseline rows are reproduced from their source papers, each on its own harness and backbone. The primary evidence is the matched-backbone paired analysis of Section~\ref{subsec:archback}. Overall accuracy (OA) is the category mean; TS-Agent's OA is the unweighted mean of the per-category scores in \cite{liu25}, computed by us, since that paper reports no overall figure.

On TimeSeriesExam both configurations exceed TS-Agent's computed overall accuracy, though the per-category profile is uneven: T-SMART leads on pattern recognition, similarity, and causality, and trails on noise. The \texttt{vision\_override} routing intervention recovers most of the anomaly deficit: the paired difference over \texttt{baseline} is $+2.7$ points overall ($n = 746$, $n_{10}/n_{01} = 58/38$, CI $[+0.1, +5.2]$, $p = 0.052$), concentrated on the anomaly-routed stratum ($+14.6$, $n = 82$, CI $[+1.2, +28.0]$), which does not survive Holm on the primary benchmark alone and is suggestive at that stratum size; the frozen out-of-distribution replication of Section~\ref{subsec:general} is the confirming evidence.
\begin{table*}[t]
\centering
\caption{TimeSeriesExam, per-category accuracy (\%). OA is the category-mean overall accuracy.}
\label{tab:tsexam}
\begin{threeparttable}
\begin{tabular}{@{}ll *{6}{S[table-format=2.1]}@{}}
\toprule
Model & Setting & {OA} & {PR} & {NU} & {AD} & {SA} & {CA} \\
\midrule
Phi-3.5 & text & 31.8 & 44 & 24 & 25 & 41 & 25 \\
Gemini-2.5 & text & 30.4 & 29 & 32 & 26 & 35 & 30 \\
Mistral-7B & text & 34.6 & 29 & 33 & 38 & 45 & 28 \\
GPT-4o & text & 38.8 & 35 & 32 & 40 & 51 & 36 \\
ChatTS & TS model & 39.0 & 37 & 39 & 36 & 53 & 30 \\
Gemini-2.5 & vision & 48.8 & 51 & 54 & 46 & 69 & 24 \\
TS-Agent & agentic, gpt-4o-mini & 60.2 & 71 & 61 & 57 & 57 & 55 \\
\midrule
\textbf{T-SMART} \texttt{baseline} & gemini-3.1 & \bfseries 65.0 & 71.8 & 59.5 & 52.8 & 70.0 & 70.8 \\
\textbf{T-SMART} \texttt{vision\_override} & gemini-3.1 & \bfseries 67.5 & 74.3 & 58.3 & 63.9 & 70.0 & 70.8 \\
\bottomrule
\end{tabular}
\begin{tablenotes}[flushleft]\footnotesize
\item PR = pattern recognition, NU = noise understanding, AD = anomaly detection, SA = similarity analysis, CA = causality analysis. OA is the unweighted mean of the five categories. Baseline rows reproduced from \cite{cai24,liu25}; T-SMART rows are this work, gemini-3.1, $n=746$. The T-SMART tool bank targets this benchmark's taxonomy (Section~\ref{subsec:branches}).
\end{tablenotes}
\end{threeparttable}
\end{table*}

On MMTS-Bench, accuracy concentrates where a question reduces to a computation the tool bank implements. With the numeric head answering free-response, T-SMART scores above ChatTS and below TS-Agent's reported figure overall, and scores highest on Base among the systems in Table~\ref{tab:mmts}. Within Base the purely tool-computed categories are at ceiling while the reasoning-heavy ones lag (noise 32.0, trend 42.0). Match is a second area of strength, consistent with the DTW pairwise design. The same table also shows the architecture's limits: on the subsets whose questions do not reduce to a registered computation, T-SMART trails TS-Agent by wide margins (InWild, Align), the measured cost of the direct-prompt fallback (Section~\ref{sec:discussion}).

\begin{table}[t]
\centering
\caption{MMTS-Bench, accuracy per subset (\%). Micro OA is the row-weighted micro-average. Base uses the numeric head on free-response.}
\label{tab:mmts}
\setlength{\tabcolsep}{4pt}
\begin{threeparttable}
\begin{tabular}{@{}l *{5}{S[table-format=2.1]}@{}}
\toprule
Model & {Micro OA} & {Base} & {InWild} & {Match} & {Align} \\
\midrule
ChatTS & 49 & 39 & 50 & 37 & 80 \\
TS-Agent & 60 & 21 & 71 & 77 & 97 \\
\textbf{T-SMART} \texttt{baseline} & \bfseries 55.6 & 72.7 & 48.3 & 59.0 & 32.9 \\
\bottomrule
\end{tabular}
\begin{tablenotes}[flushleft]\footnotesize
\item T-SMART is this work, gemini-3.1: Base $n=700$, InWild 1084, Match 400, Align 240; Micro OA is the micro-average over $n=2424$ rows.
\end{tablenotes}
\end{threeparttable}
\end{table}

\subsection{Mechanism decomposition}\label{subsec:mechanism}

\begin{table}[t]
\centering
\caption{Paired mechanism effects on identical rows (points). Top: MMTS Base, numeric head vs.\ \texttt{llm\_compute}, scored by Acc@10. Bottom: TimeSeriesExam gated subset, each visual condition vs.\ math-only.}
\label{tab:mech}
\setlength{\tabcolsep}{3pt}
\footnotesize
\begin{threeparttable}
\begin{tabular}{@{}l S[table-format=3.0] l S[table-format=+2.1] l l@{}}
\toprule
Comparison & {$n$} & {$n_{10}/n_{01}$} & {$\Delta$} & {95\% CI} & {$p$} \\
\midrule
all numerical rows & 393 & 128/12 & +29.5 & $[+24.4,+34.6]$ & $<$0.0001 \\
\quad closed-form & 328 & 104/0 & +31.7 & $[+26.8,+36.9]$ & $<$0.0001 \\
\midrule
structured vs.\ math-only & 568 & 113/69 & +7.7 & $[+3.2,+12.3]$ & 0.0014 \\
raw pixels vs.\ math-only & 568 & 119/62 & +10.0 & $[+5.5,+14.6]$ & $<$0.0001 \\
SVI cost (struct.\ $-$ raw) & 568 &  & -2.3 &  &  \\
\bottomrule
\end{tabular}
\begin{tablenotes}[flushleft]\footnotesize
\item $n_{10}/n_{01}$ = treatment-only / control-only correct; CI is the row-level percentile bootstrap of Section~\ref{subsec:eval}; $p$ is exact McNemar. The numeric rows are the two strata specified before scoring.\end{tablenotes}
\end{threeparttable}
\end{table}

The largest effect comes from deterministic computation. On the numerical rows of MMTS Base the deterministic numeric head exceeds \texttt{llm\_compute}, the same model computing the number from the raw series, by 29.5 points, rising to 31.7 on the closed-form stratum where the head wins 104 pairs and loses none (Table~\ref{tab:mech}), with the purely tool-computed categories at ceiling (basic analysis 100 vs.\ 72.0, stationarity 100 vs.\ 64.0). The wider configuration-level gap, 72.7 against 44.7 on the full 700-row subset, is not the head effect alone: because \texttt{llm\_compute} is tool-free on every row (Section~\ref{subsec:config}), 80 of its 196 aggregate rows lie on multiple-choice questions, where the difference measures the tool pipeline rather than the head. The head-specific claim is therefore the paired $+29.5$, not the $+28.0$ aggregate. The effect is not uniform: on the slope stratum the registry's estimator loses to the model ($-36.4$ points, $n = 22$, $p = 0.039$, suggestive at this size), the one quantity where the closed registry underperforms. Where the question reduces to a statistic the registry computes well, model computation from the raw values is markedly less accurate.

The model contributes as a translator (Table~\ref{tab:parser}). On a synthetic stress set of 240 questions in four layers ($n = 60$ each; Acc@10 against the tool's own answer as gold), the deterministic grammar \texttt{A1} fully recovers the compositional layer (15\% $\rightarrow$ 100\%) and most of the typo layer (0\% $\rightarrow$ 83\%). The LLM planner \texttt{A2} completes the typo layer ($+16.7$ points, $p = 0.002$) and is essential only on open paraphrase, which no deterministic parser resolves: $+75.0$ points over \texttt{A1} (CI $[+63.3, +85.0]$, $p < 0.0001$), firing on every paraphrase row and on no clean row, so the clean path is unchanged.

\begin{table}[t]
\centering
\caption{Three-parser study on the synthetic numeric stress set (Acc@10, \%; gold = the deterministic tool's own answer; $n=60$ per layer).}
\label{tab:parser}
\setlength{\tabcolsep}{5pt}
\footnotesize
\begin{tabular}{@{}l *{4}{S[table-format=3.0]}@{}}
\toprule
Layer & {\texttt{A0}} & {\texttt{A1}} & {\texttt{A2}} & {\texttt{A2} fire \%} \\
\midrule
clean & 100 & 100 & 100 & 0 \\
compositional & 15 & 100 & 100 & 0 \\
typo & 0 & 83 & 100 & 17 \\
paraphrase & 0 & 0 & 75 & 100 \\
\midrule
overall ($n{=}240$) & 29 & 71 & 94 & 29 \\
\bottomrule
\end{tabular}
\end{table}

\begin{figure}[t]
\centering
\includegraphics[width=\linewidth]{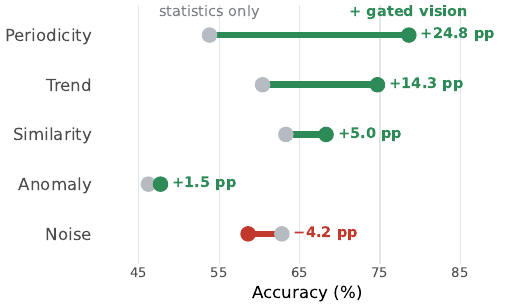}
\caption{Gated vision helps only where shape and periodicity are decisive: on gate-fired rows (gemini-3.1-flash-lite), structured vision improves periodicity ($+24.8$) and trend ($+14.3$), is near zero on anomaly, and negative on noise ($-4.2$).}
\label{fig:modality}
\end{figure}

Perception helps conditionally, and its invariant has a measured cost. On the gated TimeSeriesExam subset, structured vision beat math-only by $+8.5$ points in the initial run and $+7.7$ (95\% CI $[+3.2, +12.3]$) in the pinned regeneration, while raw pixels beat math-only by $+10.0$ in both; the cost of the Structured-Vision Invariant against raw pixels was therefore $1.5$ and $2.3$ points, with the same sign in both runs (Table~\ref{tab:mech}). One secondary contrast did not reproduce: the initial run measured far lower structured sycophancy on trend questions (0.40 versus raw's 0.73, on 10--15 discordant rows), while the regeneration finds 0.70 versus 0.73 there and comparable overall rates (0.63 structured, 0.61 raw), so we claim no sycophancy reduction for the SVI and retain the invariant on auditability grounds at its measured cost. The lift is concentrated (Figure~\ref{fig:modality}): largest on periodicity and trend, where the structured reading is also the stronger condition ($+14.3$ against raw pixels' non-significant $+9.9$), near zero on anomaly, and negative on noise, the categories a deterministic test already resolves. Vision acts as a specialist tool, most useful where shape and periodicity determine the answer.

Bounded evidence-layer correction produces no effect distinguishable from zero. Reroute-once and refine never triggered: the first trigger was never met on any of the 746 rows, and the second's candidate generator returned no usable second branch, two zeros that reflect trigger calibration on this benchmark rather than the value of rerouting or refinement, since a mechanism that never runs cannot be measured. Propose triggered on 120 of 746 rows, producing 6 corrections against 8 new errors, a paired difference of $-0.3$ points ($p = 0.79$, exact binomial on the 14 discordant rows). At this trigger rate only a large effect would be detectable, so the propose null is absence of evidence, not evidence of absence.

\subsection{Architecture versus backbone}\label{subsec:archback}
Two numbers at this backbone have different evidential status, and we report them separately. The first is within-system and paired: at gpt-4o-mini, the tool architecture with vision suppressed (\texttt{vision\_off}, isolating the tool contribution) scores 0.548 against the raw model's 0.413 (\texttt{llm\_only}) on the same 746 rows, a paired gain of 13.5 points ($n_{10}/n_{01} = 170/69$, CI $[+9.5, +17.6]$, $p < 0.0001$), significant on 5 of 6 branches under Holm. The second is cross-system and unpaired: TS-Agent's published figure at the same backbone is 0.602, 5.4 points above our loop-free system. That gap cannot be assigned to iteration, because iteration, tool set, and harness all differ at once; the loop-isolating comparison inside a single pipeline is the null of Section~\ref{subsec:nulls}. The fixed-backbone analysis establishes that T-SMART's cross-system advantage at gemini (Tables~\ref{tab:tsexam} and~\ref{tab:mmts}) is substantially a backbone effect. These gpt-4o-mini runs remain secondary evidence (Section~\ref{subsec:eval}).

\subsection{Generalization}\label{subsec:general}
The \texttt{vision\_override} routing intervention, frozen before MMTS, transfers at the mechanism level rather than as an overall gain: the anomaly stratum (91 of 1{,}865 paired rows) replicates at $+16.5$ points ($p = 0.024$), noise remains neutral, and the overall difference is $+1.0$ points and not significant. An autonomous search over vision and correction configurations, scored against a cached 504-row gpt-4o-mini anchor with per-branch paired tests under a win rule frozen before any variant ran, returned a null result: no variant met the rule, and the search ledger is included in the release.

\subsection{Schema and gate audits}\label{subsec:audits}
The schema-gate audit (1 of 686 rows off-diagonal; 0 of 388 numerical rows misread) and the learned-gate result (look-rate 76.1\% to 27.9\%, non-inferiority not met at a lower CI bound of $-2.41$ points, regression under gpt-4o-mini) are reported with their mechanisms in Sections~\ref{subsec:numhead} and~\ref{subsec:config}. Coverage on the primary benchmark is near total: the direct-prompt fallback fired on 4 of 746 rows (0.5\%) under both backbones' primary configurations, and per-comparison branch-level row counts appear in the released reports.

\section{Discussion}\label{sec:discussion}
Across the evaluated benchmarks, the decomposition is consistent. Deterministic computation accounts for most of the accuracy; the language model contributes at the language boundary; perception contributes conditionally, as a gated tool; and bounded evidence-layer correction produces no measurable gain. These claims are specific to T-SMART and the evaluated benchmarks.

Two of the paper's own results limit any broader reading. First, the computation result is scoped: the tool bank targets the same taxonomy that TimeSeriesExam and MMTS Base instantiate (Section~\ref{subsec:branches}). The InWild and Align subsets quantify performance off that taxonomy: accuracy falls to 48.3 and 32.9 while an agentic system reaches 71 and 97, because questions with no registered computation fall back to a direct prompt. The contribution is the measurement protocol and the boundaries it identifies. Second, the agency evidence is mixed. Bounded evidence-layer correction added to verified tool evidence produced nothing here, yet a published agent reports accuracy 5.4 points above ours at a matched backbone on the same closed-form benchmark (Section~\ref{subsec:archback}). Whether closed-form benchmarks can reveal the value of agency at all, and whether that gap comes from the loop or from the tools it drives, are open questions that these experiments narrow but do not answer. Two external results bound the scope of the null: chain-of-thought and multimodal fusion improve general models on MMTS-Bench \cite{yin26}, so the null concerns bounded correction atop verified tool evidence, not inference-time reasoning in general; and training-time adaptation also improves TS-QA \cite{kong25}, an option the frozen backbone excludes.

Further limitations apply. The vision estimates rest on two replications whose magnitudes moved by about one point and whose sycophancy contrast did not reproduce (Section~\ref{subsec:mechanism}). The trust gate is hand-built and permissive (Section~\ref{subsec:gate}), and its learned alternative is a suppression policy rather than a calibrated one. The parser study relies on a synthetic stress distribution, since no natural messy-numeric TS-QA benchmark exists for it. The weak-backbone analyses are secondary for the reasons of Section~\ref{subsec:eval}, and the primary results rest on a single backbone.

\section{Conclusion}\label{sec:conclusion}
We used a small, single-pass neurosymbolic system to measure where accuracy in tool-augmented TS-QA originates. On the benchmarks evaluated, deterministic computation accounts for most of the accuracy; the language model contributes at the question boundary; structured perception acts as a gated tool whose invariant costs 1.5--2.3 points against raw pixels across two replications; and three bounded evidence-layer correction loops produced no measurable gain, two of them because their triggers never fired. These are claims about one system on taxonomy-covered, closed-form benchmarks, established by paired comparisons with decision rules specified before scoring, exact tests, and multiple-comparison control. The protocol itself is designed to transfer to other systems.

\section*{Acknowledgements}
This work has been funded in part by NSF, with award numbers \#2112665, \#2112167, \#2003279, \#2120019, \#2211386, \#2052809, \#1911095 and in part by PRISM and CoCoSys, centers in JUMP 2.0, an SRC program sponsored by DARPA.

\bibliographystyle{IEEEtran}
\bibliography{refs}

\end{document}